\documentclass[letterpaper, 10 pt, journal, twoside]{IEEEtran} 
\usepackage[utf8]{inputenc}
\IEEEoverridecommandlockouts                              
\pdfoutput=1
\usepackage{mathtools}
\usepackage{graphicx}
\usepackage{booktabs}
\usepackage{amsfonts}
\usepackage{subfig}
\usepackage[linesnumbered,ruled,vlined]{algorithm2e}
\usepackage{multirow}
\usepackage{color}
\usepackage[bookmarks=false]{hyperref}
\usepackage{setspace} 

\usepackage{amssymb}

\usepackage{capt-of}

\usepackage{tikz}
\usepackage{scalerel}
\usetikzlibrary{svg.path}
\definecolor{orcidlogocol}{HTML}{A6CE39}
\tikzset{
    orcidlogo/.pic={
        \fill[orcidlogocol] 
        svg{M256,128c0,70.7-57.3,128-128,128C57.3,256,0,198.7,0,128C0,57.3,57.3,0,128,0C198.7,0,256,57.3,256,128z};
        \fill[white] svg{M86.3,186.2H70.9V79.1h15.4v48.4V186.2z}
        svg{M108.9,79.1h41.6c39.6,0,57,28.3,57,53.6c0,27.5-21.5,53.6-56.8,53.6h-41.8V79.1z 
        M124.3,172.4h24.5c34.9,0,42.9-26.5,42.9-39.7c0-21.5-13.7-39.7-43.7-39.7h-23.7V172.4z}
        svg{M88.7,56.8c0,5.5-4.5,10.1-10.1,10.1c-5.6,0-10.1-4.6-10.1-10.1c0-5.6,4.5-10.1,10.1-10.1C84.2,46.7,88.7,51.3,88.7,56.8z};
    }
}
\newcommand\orcidicon[1]{\href{https://orcid.org/#1}{\mbox{\scalerel*{
                \begin{tikzpicture}[yscale=-1,transform shape]
                \pic{orcidlogo};
                \end{tikzpicture}
            }{|}}}}

\SetKwInput{KwData}{Global params.}
\SetKw{return}{return}
\SetKw{Break}{break}

\title{Multi-goal path planning using multiple random trees}

\author{Jaroslav Jano\v{s}$^{\orcidicon{0000-0002-0467-7356}}$, Vojt\v ech Von\' asek$^{\orcidicon{0000-0001-9224-2151}}$ and Robert P\v{e}ni\v{c}ka$^{\orcidicon{0000-0001-8549-4932}}$
\thanks{Manuscript received: October 15, 2020; Revised January 18, 2021; Accepted February 20, 2021.}
\thanks{This paper was recommended for publication by Editor Nancy Amato upon evaluation of the Associate Editor and Reviewers' comments.
This work has been supported by the Czech Science Foundation (GA{\v C}R) under project No. 19-22555Y.}
\thanks{The authors are with Department of Cybernetics, Faculty of Electrical Engineering,  Czech Technical University in Prague, 
Technick\'a 2, 166 27 Prague, Czech Republic,  
{\tt janosjar@fel.cvut.cz}
}
\thanks{Digital Object Identifier (DOI): see top of this page.}
}

\def\SFFNON{NR-SFF*}
\def\rnew{r_{new}}
\def\pp{p_{q}}

\def\dist{\varrho}
\def\Tdist{\dist_{tree}}

\def\C{\mathcal{C}}
\def\CF{\mathcal{C}_{free}}

\def\OO{\mathcal{O}}

\def\Imax{I_{max}} 

\def\dist{\varrho}

\newcommand{\Cfree}[0]{\mathcal{C}_{free}}

\DeclareMathOperator*{\minimize}{\textit{minimize}}

\DeclareMathOperator*{\subjectto}{\text{s.t.}}

\begin{document}

\maketitle

\begin{abstract}
In this paper, we propose a novel sampling-based planner for multi-goal path planning among obstacles, where
the objective is to visit predefined target locations while minimizing the travel costs.
The order of visiting the targets is often achieved by  solving the Traveling Salesman Problem (TSP) or its variants.
TSP requires to define costs between the individual targets, which --- in a map with obstacles --- requires to compute
mutual paths between the targets.
These paths, found by path planning, are used both to define the costs (e.g., based on their length or time-to-traverse) and
also they define paths that are later used in the final solution.
To enable TSP finding a good-quality solution, it is necessary to find these target-to-target paths as short as possible.
We propose a sampling-based planner called Space-Filling Forest~(SFF*) that solves the part of finding collision-free paths.
SFF* uses multiple trees (forest) constructed gradually and simultaneously from the targets and attempts to find connections with other trees to form the paths. 
Unlike Rapidly-exploring Random Tree (RRT), which uses the nearest-neighbor rule for selecting
nodes for expansion, SFF* maintains an explicit list of nodes for expansion.
Individual trees are grown in a RRT* manner, i.e., with rewiring the nodes to minimize their cost.
Computational results show that SFF* provides shorter target-to-target paths than existing approaches, and consequently,
the final TSP solutions also have a lower cost.
\end{abstract}


\section{Introduction}

\enlargethispage{-4.8cm}
\noindent
\begin{picture}(0,0)
\put(0,-208){\begin{minipage}{\textwidth}
\centering
\renewcommand{\tabcolsep}{2.3pt}
{\small
\begin{tabular}{cccc}
\includegraphics[width=0.240\textwidth]{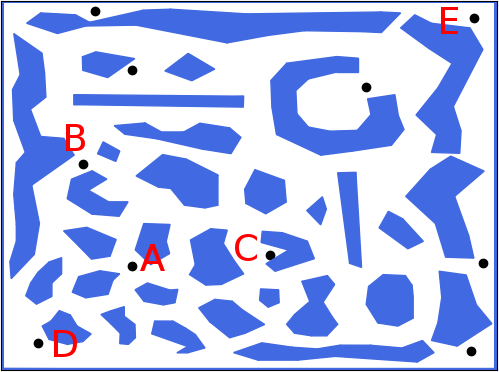} &
\includegraphics[width=0.240\textwidth]{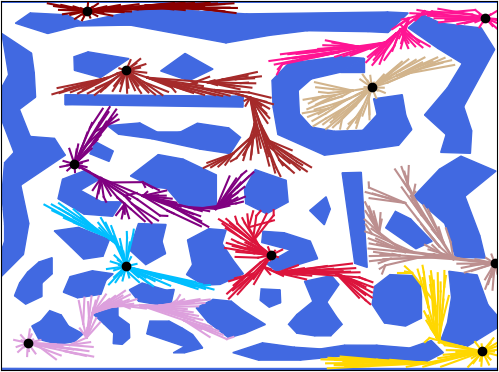} &
\includegraphics[width=0.240\textwidth]{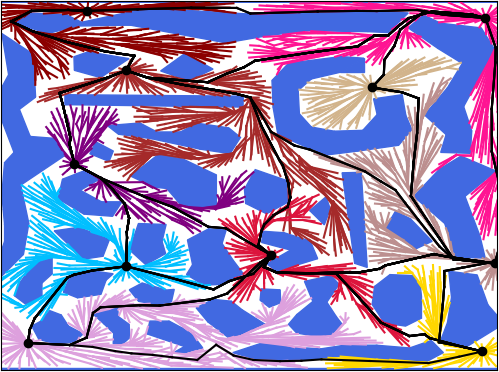} &
\includegraphics[width=0.240\textwidth]{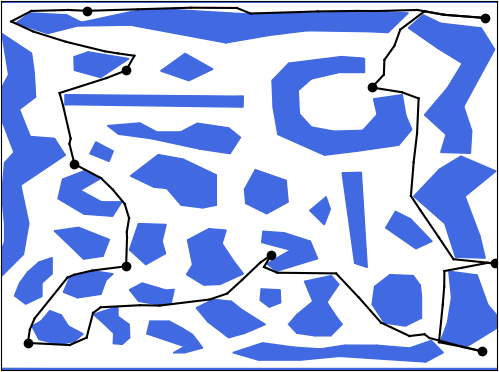} \\[-0.3em]
(a) Goals & (b) Partially grown trees  & (c) Connected trees & (d) Final path \\[-0.5em]
\end{tabular}
}
\captionof{figure}{\label{fig::intro}
\small    
Example of multi-goal path planning with target locations (black) (a).
We first build a set of trees from each goal (b) and expand them until they touch obstacles or each other (b).
Paths between the targets are found by searching the connected trees (c). 
Finally, TSP computes final sequence of goals (d).
Visualization of SFF* is available at \url{https://youtu.be/vBQVO_GP5Sc}
}
\end{minipage}}
\end{picture}%

The task of multi-goal path planning is to find a collision-free path connecting several targets~\cite{spitz2000multiple}, which is required
in data collection~\cite{faigl2014unifying,mcmahon2015autonomous}, 
active perception~\cite{best2016multi, mcmahon2015autonomous}, and
manufacturing~\cite{spitz2000multiple,saha2003planning}.
Visiting multiple goals in the shortest possible time is crucial for systems with limited operational time like flying vehicles~\cite{otto2018optimization}, e.g., for their recharging~\cite{mathew2013agraph}.

In the general version of multi-goal path planning, both the sequence of visiting the targets and also the trajectory
connecting them have to be found such that the travel cost (e.g., traveled distance or execution time) is minimized.
The classical approach, which is also considered in this paper, 
is to decouple the task to the combinatorial part (finding the sequence of targets to be visited) and to path planning part that connects
the targets in the found order.
The combinatorial phase is usually considered as an instance of Traveling Salesman Problem~(TSP) and can be solved heuristically~\cite{mcmahon2015autonomous,TSP_nearest_neighbor}.

Formulation of multi-goal path planning using the TSP requires knowledge about mutual reachability and trajectory cost between the targets.
For robots moving among obstacles, path planning between all pairs of targets is necessary to obtain the cost of their connection.
To enable TSP finding a low-cost solutions, it is furthermore desired that this target-to-target path planning provides good quality paths.

\enlargethispage{-4.8cm}

In theory, paths between all pairs of targets should be computed to define the distance matrix for TSP, but it is time-consuming.
Practically, paths between near targets are likely to be used in the final TSP sequence due to their low cost, while paths between distant targets may be ignored.
For example in Fig.~\ref{fig::intro}a, it is more useful to find paths between targets A, B, C and D as they are close to each other rather
than finding paths between D and E as the final TSP solution (Fig.~\ref{fig::intro}d) would rather connect the near targets.
This observation motivates the search proposed in this planner: instead of finding paths between all pairs of targets (to define costs
of visiting the targets for TSP), we propose to find good-quality paths only between near targets.

In this paper, we propose a novel tree-based randomized planner for finding trajectories between multiple targets concurrently (Fig.~\ref{fig::intro}a).
The planner, called Space-Filling Forest (SFF*), grows multiple trees simultaneously, starting from the given targets (Fig.~\ref{fig::intro}b).
The trees are expanded in the configuration space in a randomized manner until they approach each other or get close to an obstacle.
The nodes of two different trees are connected if they are close to each other.
This forms a roadmap where the path connecting the targets can be found (Fig.~\ref{fig::intro}c). 
The paths are then used in the subsequent TSP computation to obtain the final sequence of visiting the targets and minimizing the overall cost (Fig.~\ref{fig::intro}d).
SFF* will be released as an open-source\footnote{At \url{https://github.com/ctu-mrs/space_filling_forest_star}}. 

\section{Related Work}

Multi-goal path planning requires, besides constructing the final trajectory, also finding the order of the goals (targets) to be visited~\cite{spitz2000multiple}.
In the case of one vehicle, the problem can be considered as an instance of TSP.
Solving the combinatorial part, i.e., TSP, requires to define costs between the individual targets.
Dubins-TSP is a variant of TSP, where the robots are considered as Dubins vehicles moving in an obstacle-free environment~\cite{ny2012onthe,penicka2017dubins}. 
In Dubins-TSP, the trajectory between targets can be found analytically in a short time.

However, the approach cannot be used in the case of robots moving among obstacles, where more general planners
need to be used to obtain trajectories avoiding the obstacles.
This is formulated in the Physical TSP problem~\cite{Perez_PTSP} and used, e.g., for 
the mine countermeasure missions~\cite{mcmahon2015autonomous}.
Generally, finding paths among obstacles for robots of arbitrary shape can be solved using sampling-based planners like
Rapidly-exploring Random Tree (RRT)~\cite{lavalleRRT} and Probabilistic Roadmaps (PRM)~\cite{kavrakiForPP}.
A number of planners were derived from basic RRT and PRM, e.g., their asymptotically optimal variants RRT* and PRM*~\cite{karaman2011sampling}.
We refer to the survey~\cite{elbanhawi2014sampling} about many other variants of these planners.

A widely used approach to utilize sampling-based motion planning in multi-goal path planning is to derive
 target-to-target paths between all pairs of targets.
The final sequence of visiting the waypoints is achieved using TSP~\cite{spitz2000multiple,saha2003planning,englot2013three}.
Alternatively, other TSP-related formulations like
Watchman Routing Problem~\cite{danner2000randomized} or Vehicle Routing Problem~\cite{penicka2019pop} can be used in special cases.
The work~\cite{Placku_AUV_motion_planning_w_mission} uses a finite automaton to determine the sequence.

Finding all target-to-target trajectories is computationally demanding due to its $O(n^2)$ complexity for $n$ targets.
For scenarios with tens or a few hundreds of targets, the runtime of the TSP solution is minor
compared to the runtime of all target-to-target path planning~\cite{saha2003planning} and therefore, the speed of 
path planning is crucial.

Lazy-TSP~\cite{englot2013three} reduces the load of path planning.
For a set of targets in the environment with obstacles, the costs of their connection is simply defined
by their Euclidean distance, i.e., without considering the obstacles.
An initial tour is computed using TSP and the connection between the consecutive targets is verified by a  time-consuming RRT-based planner.
If the connection cannot be found, the TSP is iteratively refined until a valid sequence and a corresponding trajectory is found.
The number of paths computed by the RRT-based planner is therefore reduced, which also decreases the computational time.
The approach is faster than finding all target-to-target connections using the PRM approach~\cite{danner2000randomized}.
Additional speedup can be achieved using online learning methods to predict collisions of edges~\cite{kew2020batchedmotionplanning}.

The speed-up of the multi-goal planning can also be achieved by improving the underlying sampling-based 
planner, e.g., by using multiple RRT trees.
Bidirectional-RRT alternately grows two trees rooted at the start and goal, respectively, towards the random samples~\cite{kuffner2000rrt}.
One global tree (rooted at the initial configuration) and several local trees are grown in~\cite{strandberg2004augmenting}.
The random sample is first tested for connection to the global tree, and if it fails, all local trees attempt to connect to the sample.
If none of the trees can connect to the sample, a new tree is set up at the sample with a given probability.
A similar approach is presented in~\cite{clifton2008evaluating}, but in contrast to~\cite{strandberg2004augmenting}, the new tree 
is always set up at a random sample if the sample is not reachable from any other tree.
Instead of creating new trees everywhere (as in~\cite{strandberg2004augmenting,clifton2008evaluating}), the method~\cite{wang2009amulti}
establishes a new tree only if the random sample is estimated to be in a narrow passage.

Multiple RRT trees for multi-goal path planning was introduced in~\cite{Devaurs_TRRT_TSP}.
The trees are rooted at the targets, and they are selected for expansion using a round-robin, i.e., in each iteration, only
a single tree expands toward the random sample. 
If two trees approach each other close enough, they are connected (if possible), and the trajectory between the roots of the trees
is retrieved.
Then, the connected trees are merged, so they continue to grow as a single tree.
Due to the merging of the connected trees, the algorithm can provide at most one path between each pair of waypoints and no alternative
path can be found even if the number of samples increases.
Moreover, the trajectory between each pair is not optimal due to the non-optimality of the underlying planner~\cite{Devaurs_TRRT_TSP}.

In our previous work~\cite{vonasek2019space}, we proposed to grow multiple random trees and connect them at any two nodes being close enough. 
In contrary to~\cite{Devaurs_TRRT_TSP}, where the trees are merged into a single one after they approach each other, our approach~\cite{vonasek2019space} considers the connection as `virtual', i.e., the trees remain separated and grow further independently.
This results in a roadmap of trees that are connected by multiple virtual edges, so more than one path can be found between
two targets.

In this paper, we further extend our method~\cite{vonasek2019space} for multi-goal path planning in environments with obstacles; the planner
proposed in this paper is referred to as Space-Filling Forest (SFF*).
In comparison to~\cite{vonasek2019space}, we improve the quality of the paths by using the rewiring technique while growing the trees.
The growth of the trees towards other targets is boosted via a priority queue to bias finding connections between near trees.
For each tree, a list of priority queues (for every target location) is maintained; the priority queues define the priority of nodes
for the expansion.
Therefore, nodes that are believed to be close to a target location are more likely expanded.
Priority queue provides an efficient way to prioritize expansion towards promising areas and it has been already used in 
sampling-based motion planning~\cite{hernandez2019lazy}.

\section{Problem Formulation}

The multi-goal path planning problem being solved in this paper focuses on finding collision-free minimal-cost path over multiple target locations.
Such a problem includes two challenging parts.
The first one is the finding of collision-free paths with a minimal cost between all target locations.
The second one contains a combinatorial optimization problem of TSP that finds the appropriate sequence to visit the targets to minimize the overall path cost using the already found collision-free target-to-target paths.
This paper proposes the SFF* method for the first part, while the second part is solved by an existing state-of-the-art algorithm.
In the rest of this section we summarize the problem and used notation.

Let $\C$ denote the configuration space and let 
$\Cfree \subseteq \C$ is the collision-free region of the configuration space where the robot can move.
The distance between two configurations is denoted as $\dist(a,b), a,b \in \C$.

Multiple target locations are specified $R = \{r_{1}, \ldots , r_{n}\}$, $r_{i} \in \Cfree$ and need to be visited by the robot.
The sequence to visit the target locations can be described by a vector of their indexes $\Sigma = (\sigma_{1},\ldots,\sigma_{n})$, $1 \le \sigma_{i} \le n$,  $\sigma_{i} \ne \sigma_{j}$ for $i \ne j$.    
The combinatorial TSP optimization part, therefore, finds the appropriate $\Sigma$.

However, the collision-free paths connecting the targets in $R$ together with their costs have to be known before finding $\Sigma$. 
A path between targets $r_i$ and $r_j$ can be described as $\tau_{ij}: [0,1] \to \Cfree$ with $\tau_{ij}(0) = r_{i}$ and $\tau_{ij}(1) = r_{j}$.
Its cost is denoted $\dist(r_i,r_j) = \dist(\tau_{ij}) = \int_{0}^{1} |\tau_{ij}(t)| dt$.
Additionally, the shortest possible path $\tau_{ij}^{*}$ is required such that $\dist(\tau_{ij}^{*}) = \min\{\dist(\tau_{ij}) | \tau_{ij} \in \Cfree\}$.
Therefore, the necessary task is to find set 
$T^{*}=\{\tau_{ij}^{*} | i=1,\ldots,n,j=1,\ldots,n\}$ of all collision-free paths with minimal costs between the targets.

Both parts of the multi-goal path planning problem using the TSP formulation can be specified as a single optimization problem
\begin{equation} \label{op_obstacles_objective}
   \begin{split}
      \minimize_{\Sigma,T^{*}} &\sum_{i=2}^{n} \dist(r_{\sigma_{i-1}},r_{\sigma_{i}}) + \dist(r_{\sigma_{n}},r_{\sigma_{1}}) \text{, }\\
      \subjectto \text{ } \dist(r_{i},r_{j}) &= \dist(\tau_{ij}^{*}) \text{, } i=1,\ldots,n, j=1,\ldots,n\text{, }\\
      &\sigma_{i} \in \Sigma \text{, } i = 1,\ldots,n\text{.}
   \end{split}
\end{equation}

Notice that the minimality of $\tau_{ij}$ is particularly important for the paths used by $\Sigma$, which motivates the proposed SFF* to minimize mainly the path between neighboring targets as they are more often used in a TSP solution.

\section{Proposed Method}

The proposed SFF* is a sampling-based method that aims to find short paths between multiple target locations.
The method has three key features.
First, it grows multiple trees simultaneously starting from the target locations.
If two trees approach each other, their connection is tested. 
The feasible connections are considered as `virtual' edges and stored separately. 
Later, they are used to find paths between the connected trees.
Second, the nodes for expansion are explicitly defined in an open and close lists.
New nodes are added to the open list, while nodes that were difficult to expand are moved into the close list.
The algorithm primarily selects nodes for expansion from the open list, as these nodes are believed to be easily expandable.
However, the method can also draw nodes for expansion from the close list.
A node is expanded in a random direction, but expansion towards other nodes of the same tree are prohibited.
This boosts the growth towards unexplored areas and prevents each tree to grow towards itself.
Due to the usage of open/close lists, the trees are not expanded using the Voronoi-bias~\cite{lindemann2004incrementally}
(as in the case of RRT and its derivatives), yet, the trees can spread in $\Cfree$.
Third, a priority queue is used to boost the growth of each tree towards near targets.
Furthermore, the rewiring technique known from RRT*~\cite{karaman2011sampling} is used to minimize costs of the nodes.

The method is summarized in Alg.~\ref{alg::main}.
First, $n$ trees $T=\{t_1,\ldots,t_n\}$ are created and rooted in the target locations $R$ and the roots of the trees are added to the open list $O$.
In each iteration, the algorithm selects a random node (and its corresponding tree) for expansion and attempts to expand it by randomly searching
its vicinity for new collision-free nodes.
If the expansion fails, the selected node is considered as difficult to expand and it is moved to the close list $C$ 
(lines~\ref{line::remove1}--\ref{line::remove2} in Alg.~\ref{alg::main}).
The algorithm terminates after a predefined number of iterations $\Imax$ or when the open list is empty and
all target locations have been connected to a single component, i.e., when a path between each pair of targets can be found
using all created trees and considering their connection.

\newlength{\textfloatsepsave} 
\setlength{\textfloatsepsave}{\textfloatsep} 
\setlength{\textfloatsep}{0pt}

\begin{algorithm}[t]
{\small
\setstretch{0.9}
\caption{\label{alg::main}SFF}
\KwIn{%
    root nodes $R = \{r_1, \ldots, r_n\}$, priority queue bias $\pp$\;
}
\KwOut{%
path between each target $r_i,r_j \in R$
}
\hrule
$T = \{t_1,  \ldots, t_n\}$ initialize the trees at targets $r_1,\ldots, r_n$\;
$O = \{ r_1, \ldots, r_n \}$ \tcp*[r]{open list}
$C = \emptyset$ \tcp*[r]{closed list}
$E = \emptyset$ \tcp*[r]{virtual edges between trees}
initialize $Q_{i,j}, i,j = 1,\ldots,n; i \ne j$ \;
\For{$iteration = 1,\ldots, \Imax$}{
    \eIf{$O \ne \emptyset$}{
        \eIf{rand(0,1) $< \pp$}{     \nllabel{line:queue_start}
            $i = $ index of a random tree, $i \in 1,\ldots, n$ \;
            $q = $ random queue, $q \in Q_{i,j}, j=1,\ldots,n$ \;
            $e = $ best node from $q$\; \nllabel{line:queue_pop}
        }{
            $e, i = $ random node $e\in O$, index of its tree $t_i$\;\nllabel{line:rand_node_O}
        }
    }{
        $e, i = $ random node $e\in C$, index of its tree $t_i$\;\nllabel{line:rand_node_C}
    }
    \If(\tcp*[f]{Alg.~\ref{alg::expand}}){ExpandNode($e,i$)=failed {\bf and} $e \notin C$}{
      remove $e$ from open list $O$\; \nllabel{line::remove1}
      remove $e$ from all priority queues $Q_{i,j}, j=1,\ldots,n$ \; \nllabel{line::remove2}
    }
    \If{$O = \emptyset$ {\bf and} ($T+E$) forms a single component}{  \nllabel{line:single}
        \Break \tcp*[r]{we can find paths between each $r_i$ and $r_j$}
    }
}
}
\end{algorithm}

The key part of SFF* is the selection of the nodes for expansion.
Primarily, the nodes from open list are selected either randomly or by considering their distance towards
roots of other trees.
The latter case is introduced to boost the expansion towards target locations that are believed to be easily reachable.
This boosting is achieved via priority queues $Q_{i,j}, i,j = 1,\ldots,n, i \ne j$, where $i$ denotes the index of the tree 
and $j$ denotes the index of a target.
For each tree, $n-1$ queues are maintained.
The queue $Q_{i,j}$ is an ordered list of nodes of the tree $i$ according to their distance towards the target $r_j$, i.e., according
to $\dist(e, r_j), e \in t_i$.
The binary heap data structure can be used to efficiently implement the queues.

The node for expansion is selected as follows.
If the open list is not empty, a node from the open list is selected randomly with the probability $1-\pp$.
Otherwise, with the probability $\pp$, a tree is selected randomly, and then one of its queues is selected randomly as well.
From this queue, the node with the shortest distance to a particular target
is selected for expansion (lines~\ref{line:queue_start}--\ref{line:queue_pop} in Alg.~\ref{alg::main}).
If the open list is empty, the node for the expansion is selected randomly only from the close list.

\begin{figure}
\centering
{\small
\renewcommand{\tabcolsep}{1pt}
\begin{tabular}{ccc}
\includegraphics[width=0.16\textwidth]{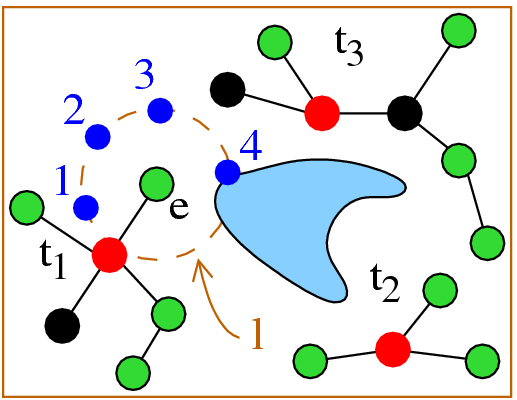} &
\includegraphics[width=0.16\textwidth]{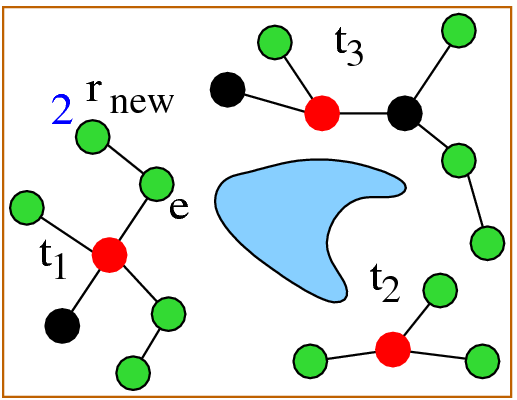} & 
\includegraphics[width=0.16\textwidth]{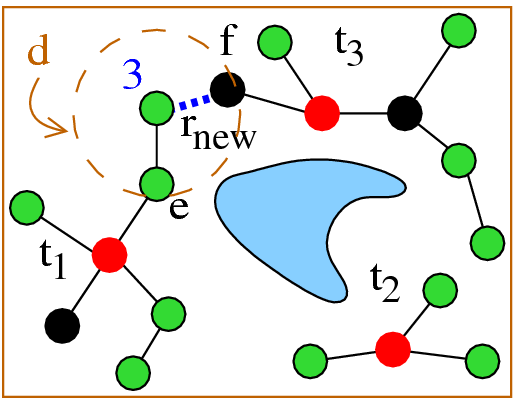} \\
(a) Sampling around      & (b) Expansion of $e$ & (c) Expansion of $e$\\
    $e$ in distance $l$  & by $\rnew$           & by $\rnew$ + connect. \\
                         &                      &  $t_1$ and $t_3$ via $\rnew$
\end{tabular}
}
\caption{\label{fig::expansion}
\small
Example of SFF* expansion for trees $t_1,t_2,t_3$ rooted at target locations (red).
The nodes in the open list are in green, the nodes in the close list are in black.
Let assume the node $e$ is going to be expanded, so its vicinity is sampled in distance $l$ from $e$ (brown) (a).
The candidate $1$ is discarded as it approaches other nodes of the tree closer than $e$.
The candidate $4$ is discarded as it is not collision-free (b).
If the candidate $\rnew = 2$ is added to the tree, it becomes member of the open list and no other action is made (b).
However, when candidate $\rnew = 3$ is added to the tree, it is already too close to the node $f \in t_3$ because
$\dist(\rnew,f) < d$. 
Therefore, the trees $t_1$ and $t_2$ are virtually connected (blue edge) via new node $\rnew$, i.e., edge $(\rnew,f)$ is added to the list
of connections $E$ (c).
\vspace{0.5em}
}
\end{figure}

The task of the expansion (Alg.~\ref{alg::expand}) is to find a new collision-free  configuration in the 
vicinity of the node being expanded such that the  tree grows toward other trees and does not grow toward itself.
We create up to $k$ samples in the distance $l$ from $e$ (node for expansion), and discard the colliding ones.
Samples that approach the same tree to the distance less than the distance to the node $e$, are also discarded, as they would
cause growing the tree toward itself.
If the expansion of a node fails, i.e., its vicinity cannot be sampled, it is removed from open list and moved to the close list.

If a new sample $\rnew$ is found, it is added to the open list, to the tree and all its queues.
The expansion process is illustrated in Fig.~\ref{fig::expansion}.
After new node $\rnew$ is added to the tree, the rewiring at $\rnew$ according to RRT* rules~\cite{karaman2011sampling} are used 
(lines~\ref{line:rew1}--\ref{line:rew2} in Alg.~\ref{alg::expand}), where $cost(e)$ denotes the length
of the path from the root of the tree to the node $e$.
The task of the rewiring is to minimize the cost of reaching the tree nodes, and consequently the cost between multiple targets when the trees are connected.

After the new node $\rnew$ is added to the tree $t_i$, the connection with other trees is checked.
The connection is possible if the distance $\Tdist(\rnew, t_j) < d$ for some tree $t_j, j \ne i$ and if this
connection is collision free (lines~\ref{line:conn1}--\ref{line:conn2} in Alg.~\ref{alg::expand}), where
$\Tdist(q,t), q\in \C, t \in T$ is the distance between a node and its nearest node in the tree $t$.
In such a case, the connecting virtual edge is remembered in the set $E$ and later used to decide if all trees form a single component or not (line~\ref{line:single} in Alg.~\ref{alg::main}).
This also allows evaluation of multiple connections between the same trees (each of them between different pairs of nodes), resulting in a different cost between targets.

\setlength{\textfloatsep}{0pt}

\begin{algorithm}[t]
{\small
\setstretch{0.9}
\caption{\label{alg::expand}ExpandNode}
\KwIn{
    index $i$ of tree  $t_i$ for expansion, node for expansion $e \in t_i$\;
}
\KwData{
    sampling distance $l$, 
    distance of two trees to be connected $d$,
    open list $O$,
    list of virtual edges $E$,
    all trees $T$ \;
}
\KwOut{
    failure or success
}
\hrule
$\rnew = \emptyset$\;
\For(\tcp*[f]{expansion around $e$}){$1,\ldots,k$}{ \nllabel{line::eat1}
    $r'=$ sample random node in the distance $l$ from $e$\;
    \If{canConnect($e,r'$) {\bf and} $\Tdist(t_i,r') > \dist(r',e)$}{ \nllabel{line:check_collisions}
        $\rnew = r'$\;
        \Break\;
    }
} \nllabel{line::eat2}
\If{$\rnew = \emptyset$}{
    \Return failure \tcp*[r]{expansion failed}
}
add $\rnew$ to tree $t_i$\;
add $\rnew$ to open list $O$\;
add $\rnew$ to all priority queues $Q_{i,j}, j=1,\ldots,n$\; 

$X_{near}$ = $k$-nearest neighbors of $\rnew$ using \cite{karaman2011sampling}, sec. 3.3.3 \; \nllabel{line:rew1}
\ForEach(\tcp*[f]{see Alg.~6 in~\cite{karaman2011sampling}}){$h\in X_{near}$}{ 
    \If{canConnect($r,h$) {\bf and} $cost(r) > cost(h) + \dist(r,h)$}{
        set $h$ as parent of $r$ and update their costs 
    }
    \If{canConnect($r,h$) {\bf and} $cost(h) > cost(r) + \dist(r,h)$}{
           set $r$ as parent of $h$ and update costs  
    } \nllabel{line:rew2}
}

\For(\tcp*[f]{connection of trees}){$t' \in T \backslash \{ t_i \}$ } { \nllabel{line:conn1}
    $f$ = find nearest node in tree $t'$ towards $\rnew$ \;
    \If{$\dist(f,\rnew) < d$ {\bf and} canConnect($f,\rnew$)}{
        \nllabel{line:check_expansion}
        add edge $(f,\rnew), f \in t', \rnew \in t_i$ to $E$\; \nllabel{line:conn2}
    }
}

\Return success\;
}
\end{algorithm}

\subsection{Discussion}

\begin{figure}
\centering
{\small
\renewcommand{\tabcolsep}{1pt}
\begin{tabular}{ccc}
\includegraphics[width=0.16\textwidth]{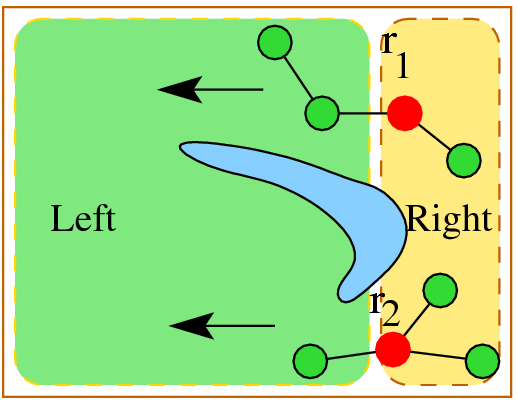} &
\includegraphics[width=0.16\textwidth]{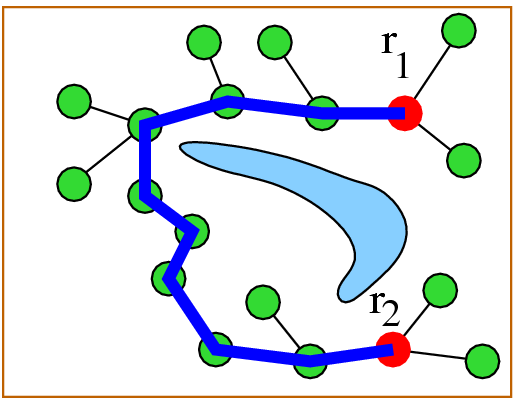} & 
\includegraphics[width=0.16\textwidth]{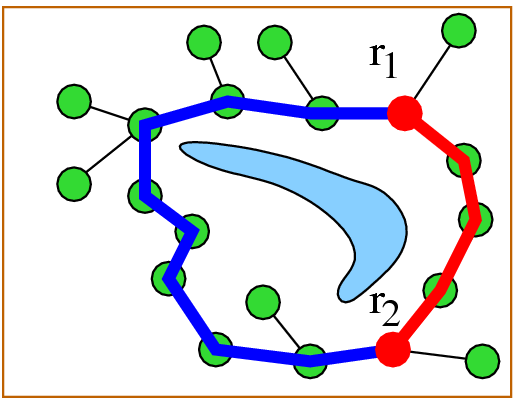} \\ 
(a)  & (b) & (c) \\
\end{tabular}
}
\caption{\label{fig::multiple}
\small
The difference between Multi-T-RRT~\cite{Devaurs_TRRT_TSP} and SFF* in a map with the target locations $r_1$ and $r_2$. 
In classic RRT (and also in~\cite{Devaurs_TRRT_TSP}), the trees are  expanded towards the random samples due to Voronoi-bias.
In the depicted scenario, more samples are generated on the left from targets than in the right (a).
Consequently, the RRT-based method, e.g.,~\cite{Devaurs_TRRT_TSP}, prefers to grow both trees towards the left zone.
The connection of these trees more likely happens also in the left zone, which results in a long path between the targets (blue) (b).
Moreover, as~\cite{Devaurs_TRRT_TSP} connects the trees only once, no other path can be found even if the number of samples is increased.
Contrary, SFF* can find multiple connections between the trees and therefore, it can also discover the shorter path (red) (c).
}
\end{figure}

As SFF* connects two trees between multiple nodes, 
we can find more paths between the same pair of target locations and considering only the shortest of them for TSP.
This feature brings an advantage over state-of-the-art planner~\cite{Devaurs_TRRT_TSP}, which can connect two trees at most once.
The comparison between SFF* and method from~\cite{Devaurs_TRRT_TSP} is illustrated in Fig.~\ref{fig::multiple}c,d.

Another advantage in comparison to~\cite{Devaurs_TRRT_TSP} is that the growth of the trees in SFF* is not driven by Voronoi-bias, but
it is maintained using the open/close lists.
Voronoi-bias boosts the growth of the RRT-based trees towards unexplored areas of the configuration space, but it can be
counteractive when connecting two near nodes located in large configuration space.
In such a situation, trees of RRT-based methods likely 
grow to the open areas of the configuration space instead of approaching each other, which may result in finding long paths (Fig.~\ref{fig::multiple}a,b,c)

The most time-consuming operations in SFF* are collision detection (CD) and the nearest-neighbor search.
The time complexity of one iteration of SFF* is
$\OO(k|t_i| \log |t_i| + kCD + REW(|t_i|) + n|t_j| \log |t_j| + nCD )$,
where the first two terms represent the nearest-neighbor search and CD when expanding the tree $t_i$ using $k$ attempts 
(lines \ref{line::eat1}--\ref{line::eat2} in Alg.~\ref{alg::expand}), where $|t_i|$ is the number of 
nodes in the tree $t_i$.
We assume that the KD-tree is used for the nearest-neighbor search.
$REW(t_i)$ is the complexity of the rewiring procedure that depends only on the size of the tree $t_i$.
The last two terms are for testing the connections between the tree $t_i$ and other trees, which also
requires to compute the nearest-neighbor and perform collision detection (lines~\ref{line:conn1}--\ref{line:conn2} in Alg.~\ref{alg::expand}).
Therefore,  the number of targets $n$ influences only the complexity of the connection of the trees.

SFF* explores the free region $\CF$ by the multiple trees whose nodes cannot be closer than $l$. 
Therefore, the number of nodes required to cover $\CF$ is given by the volume $vol(\CF)$ of $\CF$.
Let $\bar{t} \sim vol(\CF)/vol(node)$ denote the maximum number of nodes that are required
to cover $\CF$ assuming that each node has volume $vol(node)$.
The nearest-neighbor search between $\bar{t}$ nodes has time complexity $\OO(\bar{t} \log \bar{t} )$.

\begin{figure}
\centering
{
\renewcommand{\tabcolsep}{1pt}
\begin{tabular}{cc}
\includegraphics[width=0.225\textwidth]{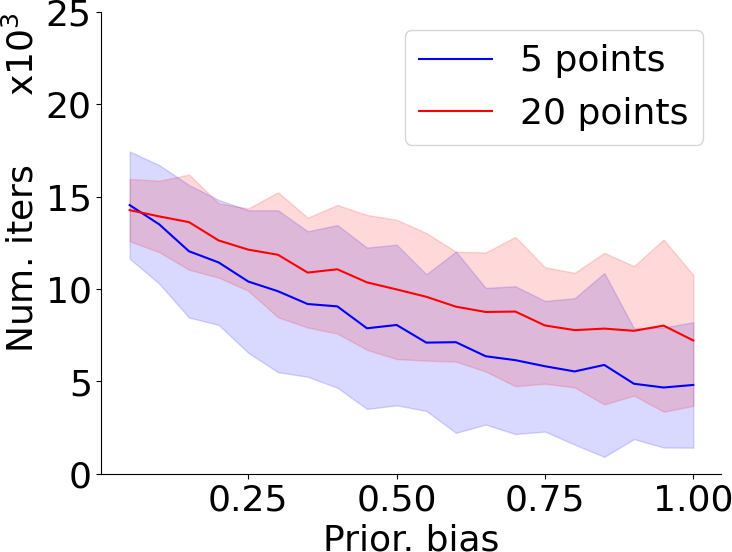} &
\includegraphics[width=0.26\textwidth]{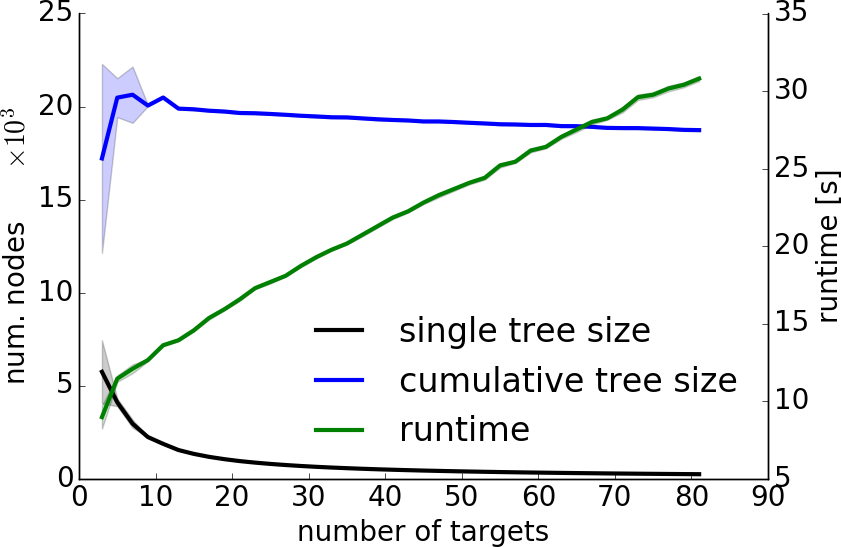} \\
(a) influence of $\pp$  & (b) influence of $n$                       
\end{tabular}
}
\caption{\label{fig::complexity}
\small
Behavior of SFF* depending on the parameter $\pp$ and with the increasing number of targets.
Graphs are made using 20 measurements in V-Dense (a) and Dense (b) maps.
}
\end{figure}

Now let's assume that $\CF$ is covered by $n$ trees and each of them has $\bar{t}/n$ nodes.
The time complexity of connecting one tree to the others 
is $\OO(n(\bar{t}/n) \log (\bar{t}/n) ) \le \OO(\bar{t} \log \bar{t})$.
Therefore, the nearest-neighbor search for connecting the trees does not depend on the number of targets, but rather
on the volume of $\CF$.
However, the expansion procedure attempts to connect actual tree $t_i$ with all other $n$ trees, which requires
$n \cdot \text{CD}$  queries.
Therefore, the overall time complexity of one expansion step can be expressed as
$\OO(k|t_i| \log |t_i| + kCD + REW(|t_i|) + \bar{t} \log \bar{t} + n CD )$, 
which increases linearly with the number of targets $n$ due to collision detection, as verified experimentally in Fig.~\ref{fig::complexity}b.

The behavior of SFF* is controlled via several parameters. 
The parameter $l$ specifies the size of the neighborhood where new samples are generated around a node being expanded.
With increasing $l$, the tree grows faster (spreading more with few iterations), but it can miss the entrance to more difficult
regions, e.g. to narrow passages. 
Therefore, the proper value depends mainly on the map, and we recommend to start with $l$ according to the width of the expected
narrow passages in the environments.
The number of samples per expansion is determined by $k$. 
Low values speed up the expansion step, but the performance may be decreased in the narrow passages.
We recommend to set up $k\approx 10$.
Two trees are virtually connected if they approach each other to the distance $d$. 
Larger $d$ means that more connection tests will be performed, which increases the runtime as these connections
necessarily rely on collision detection.
We recommend to set $d$ as a small multiple of $l$, e.g. as $d=2\cdot l$.
With the higher probability of using the priority queues, $\pp$, SFF* prefers to connect near trees in early stages of the method, which also decreases the number of
iterations required to find all target-to-target paths (Fig.~\ref{fig::complexity}a).
Our experiments show that $\pp=0.9$--$0.95$ yield the best results.

SFF* can also be used in other planning problems that  require search in the configuration space.
However, the design of SFF* assumes that several trees are grown simultaneously and that discovering multiple connections between the trees
brings an advantage in the given task.
Applying SFF* on a planning problem with a single start/goal is possible, but in such a case, the method reduces to RRT*, and
the proposed features, e.g., priority heap, are not in use.
SFF* can principally search high-dimensional configuration space.
It only requires to employ a suitable nearest-neighbor data structure.
For non-holonomic systems, a local planner allowing exact connection of two near configuration is required.
These requirements are common also for other sampling-based planners.

\section{Results}

The SFF* was compared with the Multi-T-RRT planner~\cite{Devaurs_TRRT_TSP}, Lazy-TSP~\cite{englot2013three},
and with the Simple-SFF~\cite{vonasek2019space}.
The contribution of the tree rewiring was further examined by using \SFFNON\ version, which is SFF*, where
rewiring is disabled (i.e., lines \ref{line:rew1}--\ref{line:rew2} in Alg.~\ref{alg::expand} are disabled).
All methods were implemented in C++, and the experiments were performed on eight 16-core AMD~Opteron~6376 2.3~GHz processors with 48~GB of RAM.

\subsection{Performance in 2D workspace}

In the first set of experiments, paths between 5, 10 and 20 targets has to be found for 
a holonomic hexagonal robot of radius $10$ units.
Three scenarios were considered: Dense, V-Dense and Triangles with size $2000\times2000$ units (Fig.~\ref{fig::tsp2d}).

On average, finding the multi-goal paths on problems with 20 targets took
$0.58~\mathrm{s}$ for Multi-T-RRT,    
$13.49~\mathrm{s}$ for Simple-SFF,
$13.83~\mathrm{s}$ for \SFFNON, and $16.19~\mathrm{s}$ for SFF*.
Lazy-TSP is considerable slower, which is caused by the RRT* planner that is internally used in Lazy-TSP to evaluate
target-to-target distances.
On the problems with 5 and 10 targets, Lazy-TSP took in average $18~\mathrm{s}$, 
but $>800~\mathrm{s}$ was required to solve problems with 20 targets.

The result of each planner is a roadmap in which target-to-target paths were found using Dijkstra's algorithm, and
costs of these paths were used in TSP.
TSP was calculated using the state-of-the-art TSP Concorde solver~\cite{concorde_solver}.
The runtimes of TSP are, in our scenarios with few tens of nodes, negligible compared to the runtime of the planners.
Namely, TSP with 5 targets is solved in $\sim20$~ms, $\sim26$~ms for 10 targets, and $\sim125$~ms for 20 targets.

From the runtime point of view, Multi-T-RRT outperforms the proposed SFF*.
However, as will be demonstrated in the following text, the proposed SFF* (and also Simple-SFF and \SFFNON) provides
significantly better paths than Multi-T-RRT.
A practitioner may want to know whether running Multi-T-RRT repeatedly and using the best solution (i.e., a solution with
the shortest paths between the targets) can yield similar results as SFF*, but with a shorter computational time.
We investigated this alternative by considering another method called Multi-T-RRT-20, which represents the best 
results out of 20 trials of Multi-T-RRT.
We decided to use 20 trials as a single run of Multi-T-RRT is approximately 28 times faster than SFF*, therefore 
running 20 times Multi-T-RRT would still result in a faster planner than SFF*.

SFF* and NR-SFF* was run with $\pp=0.95$, $d=50$ units and $l=40$ units and $k=12$.
Collision detection was called $3(n+k)$ times per iteration of SFF*, because each iteration results in $n+k$ edges that are tested
for collisions at three points (start/middle/end).
Multi-T-RRT was run in a similar manner: the expansion step was $l=40$ units and the distance for connecting two trees
was $d=50$ units.
All methods were terminated after $\Imax = 100 \times 10^3$ iterations.

\subsection{Target-to-target distance comparison}

We first compare the costs of the paths between the targets as follows.
After each algorithm finishes, we find all target-to-target paths.
The cumulative cost of these paths is then used for the comparison.
For each map and the number of target locations, each algorithm was run $10^4$ times.

\begin{figure}[!t]
\begin{center}
{\small
\def\sirka{0.24}
\def\vyska{0.15}
\renewcommand{\tabcolsep}{0.5pt}
\begin{tabular}{cc}
\includegraphics[width=\sirka\textwidth,height=\vyska\textwidth]{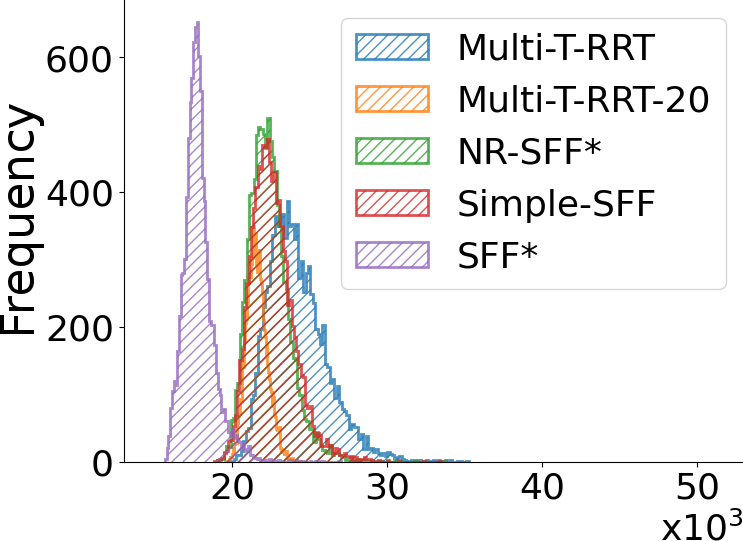} &
\includegraphics[width=\sirka\textwidth,height=\vyska\textwidth]{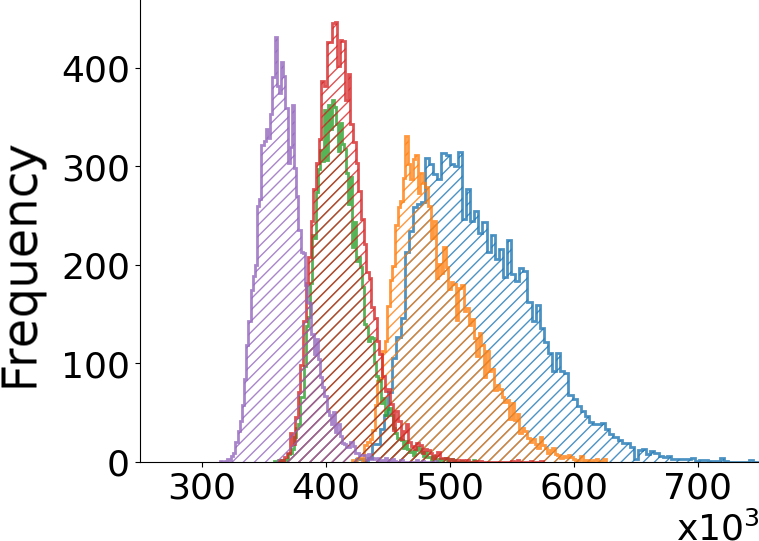} \\
(a) Dense, 10 targets & (b) Dense, 20 targets \\
\includegraphics[width=\sirka\textwidth,height=\vyska\textwidth]{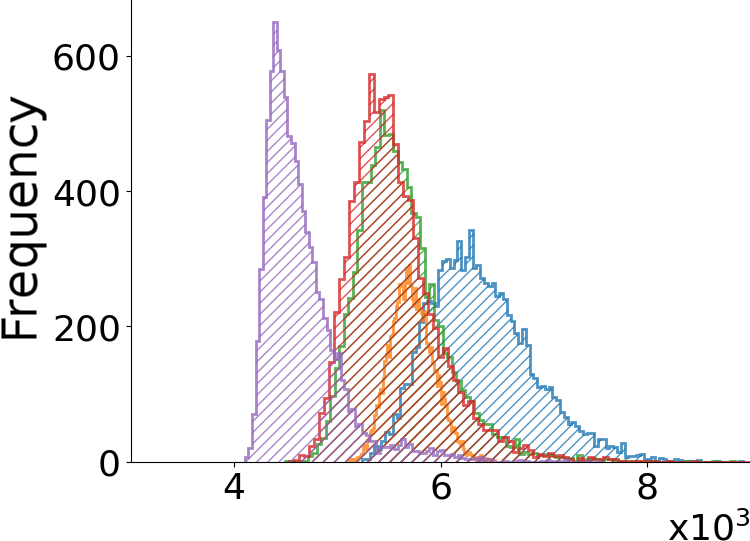} &
\includegraphics[width=\sirka\textwidth,height=\vyska\textwidth]{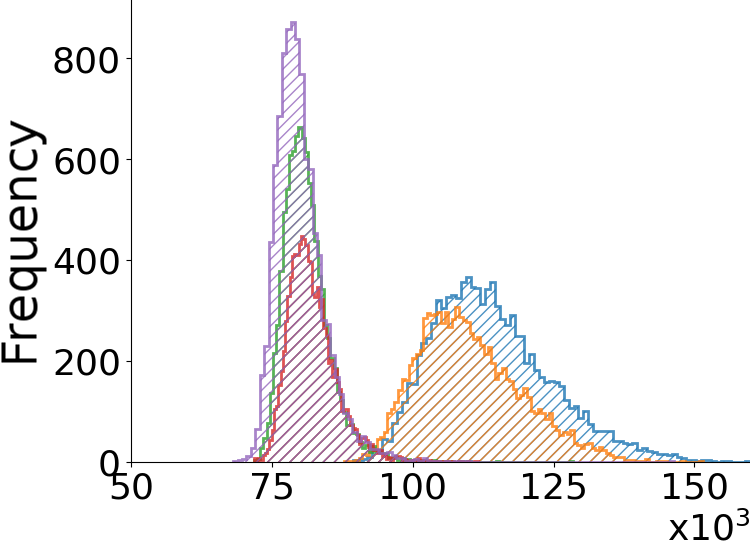} \\[-0.5em]
(c) Triangles, 10 targets & (d) Triangles, 20 targets \\[0.5em]                       
\includegraphics[width=\sirka\textwidth,height=\vyska\textwidth]{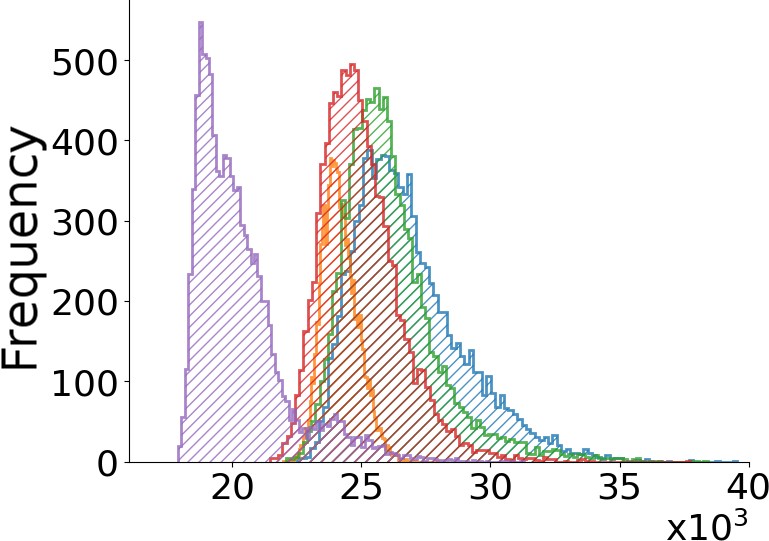} &
\includegraphics[width=\sirka\textwidth,height=\vyska\textwidth]{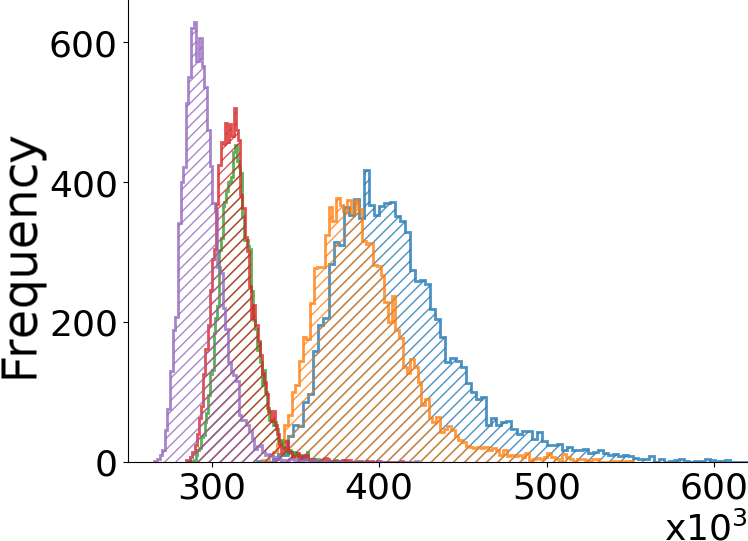}  \\[-0.5em]
(d) V-Dense, 10 targets & (f) V-Dense, 20 targets \\
\end{tabular}
}
\caption{\label{fig::cumulative}
\small  Histograms of cumulative costs of target-to-target paths, the cost path (horizontal axis) is the map units.
}
\end{center}
\end{figure}

The results are depicted in Fig.~\ref{fig::cumulative} in the form of a histogram of the cumulative costs (in units).
In all tested cases, SFF* outperformed other methods because it provides smaller cumulative costs, i.e., shorter paths,
than other methods.
The planner Multi-T-RRT provides the highest cumulative costs. 
Repeated runs of Multi-T-RRT with selecting the best solution out of 20 trials (algorithm Multi-T-RRT-20) decreases
the cumulative costs, but not significantly, so even Multi-T-RRT-20 is outperformed by SFF*.

Multi-T-RRT is also outperformed by the Simple-SFF (which uses no rewiring and no priority heap but connects
the trees multiple times).
This indicates that connecting the trees multiple times already improves the quality of the solutions.
Contrary, Multi-T-RRT connects the trees only once, which results in only one path connecting each target location.
Due to the stochastic behavior of Multi-T-RRT, this connection varies between different runs, which also results
in a higher deviation of the cumulative costs. 
Contrary, the deviation of SFF*, Simple SFF and \SFFNON\ is smaller.
The comparison between SFF and \SFFNON\ shows the positive effect of the rewiring procedure: 
SFF* provides better results (lower values of the cumulative costs) than \SFFNON.
Lazy-TSP is not included in Fig.~\ref{fig::cumulative} as it does not compute all target-to-target paths.

\def\NOTEA{\textbf{$^\mathbf{a}$}}
\def\NOTEB{\textbf{$^\mathbf{b}$}}
\def\DIFF{$\ne$}
\def\SAME{$=$}

\begin{table*}[!htb]
\centering
{\scriptsize
\renewcommand{\arraystretch}{0.85}
\renewcommand{\tabcolsep}{4pt}
\caption{\label{tab::tsp}
\small
Comparison of the planners using costs of TSP solutions. 
The columns `TSP' and `Iterations' are in format mean $|$ std. dev. 
The results are based on 100 measurements.
`TSP best' is the smallest achieved TSP cost.
TSP costs are computed using Concorde~\cite{concorde_solver} solver.
Computing TSP using LKH~\cite{helsgaun2017extension} led to same results in all cases except ones 
denoted \NOTEA, where the costs of LKH solution is by 0.21~\% smaller than the solution of Concorde; 
and in case \NOTEB, where the cost of LKH solution is by 0.33~\% smaller than the solution of Concorde.
The marks \DIFF\ and \SAME\ denote if the cost of SFF* solution is different, or same, as solution of Lazy-TSP using t-test and $\alpha=0.05$. 
}
\begin{tabular}{lccccccccc}
\toprule
 Num. of goals: & \multicolumn{3}{c}{ \bf 5 } & \multicolumn{3}{c}{ \bf 10 } & \multicolumn{3}{c}{ \bf 20 } \\ 
 \cmidrule(lr){2-4}  \cmidrule(lr){5-7}  \cmidrule(lr){8-10} 
 & {\bf TSP } & {\bf TSP best} &  {\bf Iterations}  & {\bf TSP } & {\bf TSP best} &  {\bf Iterations}  & {\bf TSP } & {\bf TSP best} &  {\bf Iterations} \\ 
 & $\times 10^3$ & $\times 10^3$ & $\times 10^3$  & $\times 10^3$ & $\times 10^3$ & $\times 10^3$  & $\times 10^3$ & $\times 10^3$ & $\times 10^3$ \\ 
\midrule
\multicolumn{10}{l}{ \bf Dense} \\ 
{ \bf SFF*} 
 & { \bf  7.00 $|$ 0.53 }  & \textbf{  \textit{ 6.46 }  }  &  5.60 $|$ 0.32  &  12.63 $|$ 1.34\NOTEA  & \textit{ 9.99 }  &  10.72 $|$ 10.74  &  15.18 $|$ 1.23 \NOTEA & \textit{ 12.25 }  &  12.25 $|$ 12.53 \\ 
{ \bf \SFFNON} 
 &  9.09 $|$ 0.61  & \textit{ 7.60 }  &  6.21 $|$ 0.36  &  15.04 $|$ 1.50\NOTEA  & \textit{ 11.82 }  &  10.93 $|$ 10.30  &  17.44 $|$ 1.33 \NOTEA  & \textit{ 13.87 }  &  12.29 $|$ 11.97 \\ 
{ \bf Simple-SFF} 
 &  9.33 $|$ 0.74  & \textit{ 7.64 }  &  6.01 $|$ 0.35  &  15.00 $|$ 1.45\NOTEA  & \textit{ 11.87 }  &  11.40 $|$ 10.94  &  17.60 $|$ 1.40\NOTEB  & \textit{ 14.30 }  &  13.76 $|$ 13.49 \\ 
{ \bf Multi-T-RRT} 
 &  10.85 $|$ 0.77  & \textit{ 8.93 }  &  0.47 $|$ 0.14  &  20.03 $|$ 1.38  & \textit{ 15.65 }  &  2.10 $|$ 0.60  &  25.33 $|$ 1.38  & \textit{ 21.46 }  &  2.29 $|$ 0.61 \\ 
{ \bf Multi-T-RRT-20} 
 &  9.66 $|$ 0.23  & \textit{ 8.85 }  &  0.40 $|$ 0.11  &  17.76 $|$ 0.54  & \textit{ 15.59 }  &  1.91 $|$ 0.57  &  23.08 $|$ 0.52  & \textit{ 20.81 }  &  2.20 $|$ 0.59 \\ 
{ \bf Lazy TSP} 
 &  7.18 $|$ 0.40 \DIFF & \textit{ 6.59 }  &  20.99 $|$ 8.08  & { \bf  10.30 $|$ 0.50 } \DIFF & \textbf{  \textit{ 9.34 }  }  &  84.53 $|$ 28.47  & { \bf  12.73 $|$ 0.39 } \DIFF & \textbf{  \textit{ 11.95 }  }  &  125.94 $|$ 32.99 \\ 
\midrule
\multicolumn{10}{l}{ \bf Triangles} \\ 
{ \bf SFF*} 
 &  1.82 $|$ 0.19  & \textbf{  \textit{ 1.59 }  }  &  2.84 $|$ 2.99  &  2.42 $|$ 0.18  & \textbf{  \textit{ 2.07 }  }  &  2.61 $|$ 1.28  &  3.22 $|$ 0.20  & \textbf{  \textit{ 2.65 }  }  &  2.45 $|$ 1.25 \\ 
{ \bf \SFFNON} 
 &  2.25 $|$ 0.22  & \textit{ 1.74 }  &  2.97 $|$ 2.13  &  2.74 $|$ 0.19  & \textit{ 2.24 }  &  2.76 $|$ 1.95  &  3.36 $|$ 0.21  & \textit{ 2.85 }  &  2.53 $|$ 1.44 \\ 
{ \bf Simple-SFF} 
 &  2.23 $|$ 0.23  & \textit{ 1.73 }  &  2.90 $|$ 2.41  &  2.75 $|$ 0.22  & \textit{ 2.23 }  &  2.69 $|$ 1.03  &  3.42 $|$ 0.23  & \textit{ 2.85 }  &  2.54 $|$ 1.48 \\ 
{ \bf Multi-T-RRT} 
 &  2.78 $|$ 0.21  & \textit{ 2.18 }  &  0.10 $|$ 0.03  &  4.07 $|$ 0.24  & \textit{ 3.38 }  &  0.13 $|$ 0.03  &  5.42 $|$ 0.24  & \textit{ 4.65 }  &  0.14 $|$ 0.03 \\ 
{ \bf Multi-T-RRT-20} 
 &  2.42 $|$ 0.08  & \textit{ 2.15 }  &  0.09 $|$ 0.02  &  3.66 $|$ 0.10  & \textit{ 3.31 }  &  0.12 $|$ 0.03  &  5.00 $|$ 0.11  & \textit{ 4.48 }  &  0.13 $|$ 0.02 \\ 
{ \bf Lazy TSP} 
 & { \bf  1.81 $|$ 0.09 } \SAME & \textit{ 1.64 }  &  1.74 $|$ 0.79  & { \bf  2.35 $|$ 0.10 } \DIFF & \textit{ 2.11 }  &  3.52 $|$ 7.12  & { \bf  3.00 $|$ 0.10 } \DIFF & \textit{ 2.75 }  &  3.29 $|$ 0.83 \\ 
\midrule
\multicolumn{10}{l}{ \bf V-Dense} \\ 
{ \bf SFF*} 
 &  8.55 $|$ 0.80  & \textbf{  \textit{ 7.61 }  }  &  5.49 $|$ 0.60  & { \bf  9.17 $|$ 0.50 }  & \textbf{  \textit{ 8.46 }  }  &  5.58 $|$ 0.51  &  12.35 $|$ 0.56  & \textbf{  \textit{ 11.13 }  }  &  5.67 $|$ 0.99 \\ 
{ \bf \SFFNON} 
 &  11.37 $|$ 0.93  & \textit{ 9.33 }  &  6.10 $|$ 0.65  &  11.37 $|$ 0.58  & \textit{ 9.92 }  &  6.09 $|$ 0.81  &  14.61 $|$ 0.64  & \textit{ 12.63 }  &  6.10 $|$ 1.23 \\ 
{ \bf Simple-SFF} 
 &  11.00 $|$ 0.94  & \textit{ 9.00 }  &  5.89 $|$ 1.03  &  11.39 $|$ 0.65  & \textit{ 9.85 }  &  5.94 $|$ 1.09  &  14.51 $|$ 0.71  & \textit{ 12.67 }  &  6.00 $|$ 1.33 \\ 
{ \bf Multi-T-RRT} 
 &  12.42 $|$ 0.70  & \textit{ 10.67 }  &  0.45 $|$ 0.21  &  15.99 $|$ 0.89  & \textit{ 13.51 }  &  0.48 $|$ 0.18  &  21.64 $|$ 0.84  & \textit{ 19.14 }  &  0.49 $|$ 0.15 \\ 
{ \bf Multi-T-RRT-20} 
 &  11.36 $|$ 0.21  & \textit{ 10.41 }  &  0.36 $|$ 0.16  &  14.58 $|$ 0.31  & \textit{ 13.26 }  &  0.44 $|$ 0.16  &  20.19 $|$ 0.36  & \textit{ 18.42 }  &  0.44 $|$ 0.13 \\ 
{ \bf Lazy TSP} 
 & { \bf  8.49 $|$ 0.29 } \SAME & \textit{ 7.81 }  &  14.82 $|$ 5.76  &  9.50 $|$ 0.38 \DIFF & \textit{ 8.81 }  &  34.12 $|$ 9.62  & { \bf  12.28 $|$ 0.36 } \SAME & \textit{ 11.30 }  &  51.39 $|$ 12.62 \\ 
\bottomrule
\end{tabular}

}
\end{table*}

\subsection{Multi-goal paths }

\begin{figure}
\vspace{-1.0em}
{\small
\renewcommand{\tabcolsep}{1pt}
\begin{tabular}{cc}
\includegraphics[width=0.24\textwidth]{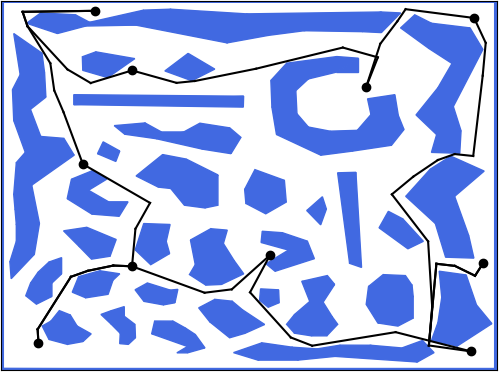}  &
\includegraphics[width=0.24\textwidth]{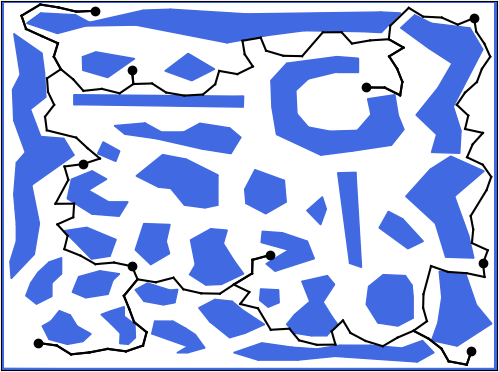} \\
(a ) Dense, SFF* & (b) Dense, \SFFNON \\
\includegraphics[width=0.23\textwidth]{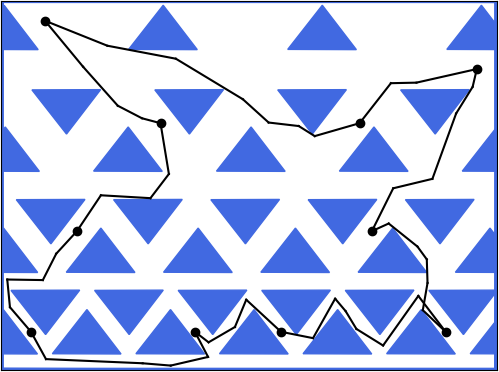} &
\includegraphics[width=0.23\textwidth]{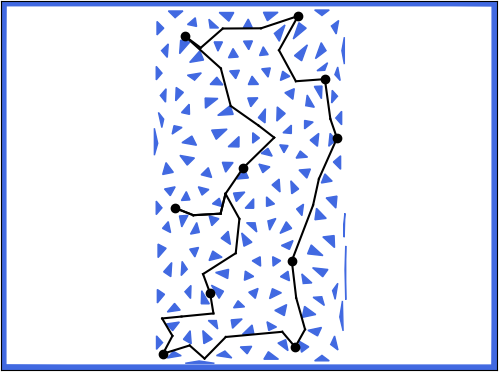}  \\
(c) V-Dense, SFF*  & (d) Triangles, SFF*
\end{tabular}
}
\caption{\label{fig::tsp2d}\small  Final TSP solution on the V-Dense map (a) and on the Triangles map (b).
}
\vspace{0.5em}
\end{figure}

Example of the final multi-goal tours are depicted in Fig.~\ref{fig::tsp2d}.
The progress of SFF* trees is depicted in Fig.~\ref{fig::intro}.
The results are summarized in Tab.~\ref{tab::tsp}, where the column `TSP' is the total length of the final paths (in units),
the column `Iterations' denotes how many sampling iterations were needed to connect all targets to a single component.
The shortest paths are provided by SFF* and Lazy-TSP.
The cases where Lazy-TSP and SFF* provided statistically same results (using t-test with $\alpha=0.05$) are marked `\SAME' in the table.
The biggest advantage of the proposed SFF* over Lazy-TSP is the number of iterations required to find the solution: Lazy-TSP requires
5--10$\times$ more iterations than the SFF*. 
Consequently, also the runtime of Lazy-TSP is significantly higher than the runtime of SFF*.
This enables SFF* to be run repeatedly and using the best result with the minimum TSP cost.
In this case, the best result provided by SFF* is better than the best result provided by Lazy-TSP (column `TSP best' in Tab.~\ref{tab::tsp}).

The cost of TSP solutions obtained using paths from Multi-T-RRT is from 45.26~\% to 75.23~\% higher than the costs
of solutions achieved using SFF*. 
When repeated runs of Multi-T-RRT are used, the cost of TSP solutions is from 32.54~\% to 63.48~\% higher than the cost of
TSP solutions obtained using SFF*.
This indicates that running a planner providing low-quality paths repeatedly, albeit it is faster, does not necessarily improve the cost
of the final solution.
We have also examined the influence of TSP solver.
Besides Concorde, that was used to compute TSP solutions in Tab.~\ref{tab::tsp}, also LKH~\cite{helsgaun2017extension} was used.
In our scenarios with few tens of targets, Concorde and LKH provided the same results, except few cases that are marked in Tab.~\ref{tab::tsp}.

\subsection{Performance in 3D workspace}

Multi-goal path planning in 6D configuration space (3D robot in 3D workspace) was tested using three scenarios (Fig.~\ref{fig::map3d}) and compared to Multi-T-RRT.
The size of the workspace is $100 \times 100 \times 100$ units and robot was 10 units long with radius 3 units.
In average, SFF* takes 10 minutes on the scenarios with 20 targets and Multi-T-RRT takes 5 minutes.
The speed of Lazy-TSP was too high ($>50$ minutes per trial for more than 10 and 20 targets), and therefore, Lazy-TSP was not tested in this case.
The results are summarized in Tab.~\ref{tab::tsp3d}. 
SFF* utilized all allowed iterations ($\Imax=100 \times 10^3$) but still found multiple connections between the trees.
The reason for so many iterations is the growth of the trees outwards from the obstacles (towards the open space), not the inability
to connect individual trees.
This is confirmed by the quality of TSP paths, which is lower for SFF* and higher for Multi-T-RRT.

\begin{table*}[!htb]
\centering
{\scriptsize
\renewcommand{\arraystretch}{0.85}
\renewcommand{\tabcolsep}{4pt}
\caption{\label{tab::tsp3d}
\small
Comparison of the planners using costs of TSP solutions. 
The columns `TSP' and `Iterations' are in format mean $|$ std. dev. 
The results are based on 100 measurements.
`TSP best' is the smallest achieved TSP cost.
TSP costs are computed using Concorde~\cite{concorde_solver} solver and the same TSP costs were obtained
also using LKH solver~\cite{helsgaun2017extension}.
}
{
\begin{tabular}{lccccccccc}
\toprule
 Num. of goals: & \multicolumn{3}{c}{ \bf 5 } & \multicolumn{3}{c}{ \bf 10 } & \multicolumn{3}{c}{ \bf 20 } \\ 
 \cmidrule(lr){2-4}  \cmidrule(lr){5-7}  \cmidrule(lr){8-10} 
 & {\bf TSP } & {\bf TSP best} &  {\bf Iterations}  & {\bf TSP } & {\bf TSP best} &  {\bf Iterations}  & {\bf TSP } & {\bf TSP best} &  {\bf Iterations} \\ 
 & $\times 10^3$ & $\times 10^3$ & $\times 10^3$  & $\times 10^3$ & $\times 10^3$ & $\times 10^3$  & $\times 10^3$ & $\times 10^3$ & $\times 10^3$ \\ 
\midrule
\multicolumn{10}{l}{ \bf Dense 3D} \\ 
{ \bf Multi-T-RRT} 
 &  14.99 $|$ 1.81  & \textit{ 11.05 }  &  1.44 $|$ 0.52  &  29.20 $|$ 2.35  & \textit{ 23.80 }  &  2.15 $|$ 0.53  &  46.58 $|$ 2.96  & \textit{ 39.82 }  &  2.45 $|$ 0.50 \\ 
{ \bf SFF*} 
 & { \bf  6.79 $|$ 0.14 }  & \textbf{  \textit{ 6.50 }  }  &  34.05 $|$ 0.39  & { \bf  10.63 $|$ 0.50 }  & \textbf{  \textit{ 9.96 }  }  &  32.77 $|$ 0.33  & { \bf  14.29 $|$ 0.49 }  & \textbf{  \textit{ 13.48 }  }  &  32.64 $|$ 2.09 \\ 
\midrule
\multicolumn{10}{l}{ \bf Building} \\ 
{ \bf Multi-T-RRT} 
 &  0.16 $|$ 0.01  & \textit{ 0.12 }  &  27.74 $|$ 13.88  &  0.27 $|$ 0.02  & \textit{ 0.22 }  &  18.28 $|$ 10.00  &  0.40 $|$ 0.02  & \textit{ 0.34 }  &  19.27 $|$ 9.64 \\ 
{ \bf SFF*} 
 & { \bf  0.06 $|$ 0.00 }  & \textbf{  \textit{ 0.06 }  }  &  100.00 $|$ 0.00  & { \bf  0.09 $|$ 0.00 }  & \textbf{  \textit{ 0.08 }  }  &  100.00 $|$ 0.00  & { \bf  0.13 $|$ 0.00 }  & \textbf{  \textit{ 0.12 }  }  &  100.00 $|$ 0.00 \\ 
\midrule
\multicolumn{10}{l}{ \bf Triangles 3D} \\ 
{ \bf Multi-T-RRT} 
 &  0.11 $|$ 0.01  & \textit{ 0.08 }  &  3.82 $|$ 1.21  &  0.20 $|$ 0.02  & \textit{ 0.16 }  &  5.07 $|$ 1.31  &  0.31 $|$ 0.02  & \textit{ 0.25 }  &  6.04 $|$ 1.33 \\ 
{ \bf SFF*} 
 & { \bf  0.04 $|$ 0.00 }  & \textbf{  \textit{ 0.04 }  }  &  100.00 $|$ 0.00  & { \bf  0.06 $|$ 0.00 }  & \textbf{  \textit{ 0.06 }  }  &  100.00 $|$ 0.00  & { \bf  0.09 $|$ 0.00 }  & \textbf{  \textit{ 0.08 }  }  &  100.00 $|$ 0.00 \\ 
\bottomrule
\end{tabular}

}
}
\end{table*}

\begin{figure}
\vspace{-1.0em}
\centering
{\small
\renewcommand{\tabcolsep}{1pt}
\begin{tabular}{ccc}
\includegraphics[width=0.19\textwidth]{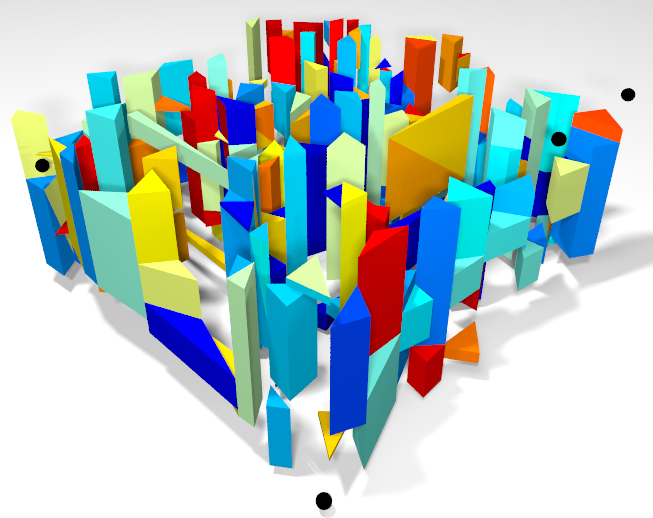} &
\includegraphics[width=0.15\textwidth]{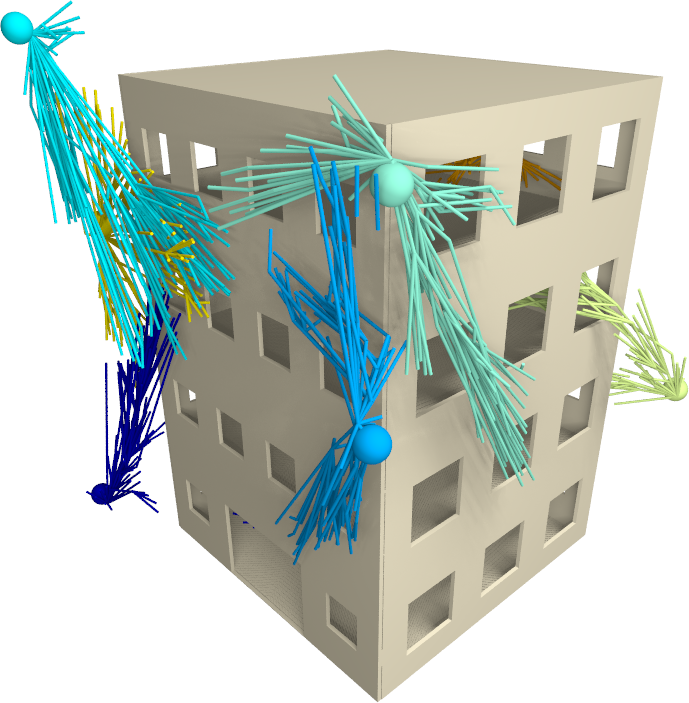} &
\includegraphics[width=0.14\textwidth]{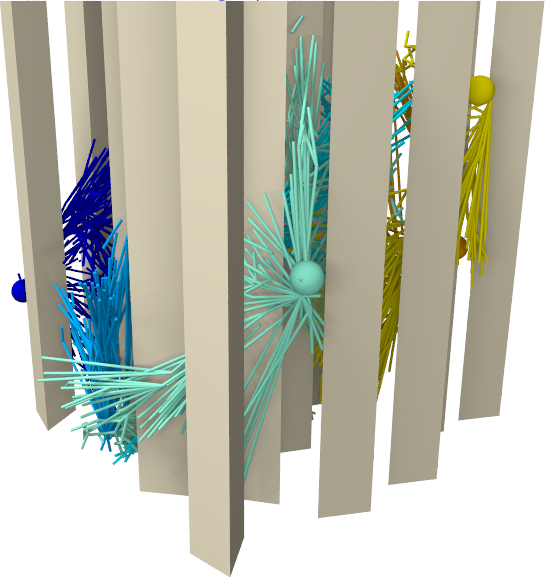} \\
(a) Dense 3D & (b) Building & (c) Triangles 3D \\                       
\end{tabular}
}
\caption{\label{fig::map3d}\small 3D workspace with partially grown SFF* trees. Spheres
    denote the target locations.
    \vspace{0.5em}
}
\end{figure}

\section{Conclusion}

This paper focused on multi-goal path planning, where the task is to visit several target locations in an environment with obstacles.
The order of visiting the targets is obtained by solving the related Traveling Salesman Problem.
The core of the paper is the novel path planner called Space-Filling Forest (SFF*) that finds high-quality paths between the individual targets.
SFF* builds multiple trees in the configuration space starting from the target locations.
The trees are grown in the free space until they approach each other or their growth is locally stopped by an obstacle.
Multiple connections are found between the trees that approached each other to a predefined distance.
Paths between the targets are found in the connected trees.
To find  good-quality paths, the trees are grown in the RRT* manner, i.e., with rewiring the nodes during the growth.
The experiments have shown superior performance in comparison to state-of-the-art methods.

\bibliographystyle{plain}
\bibliography{paper}

\end{document}